\documentclass{article}

\usepackage{ifthen}
\newboolean{showcomments}
\setboolean{showcomments}{true}

\usepackage{microtype}
\usepackage{graphicx}
\usepackage{booktabs} 
\usepackage{listings}
\usepackage{amsmath}
\usepackage{algorithm}
\usepackage{algorithmic}

\usepackage{graphicx}
\usepackage{grffile}
\usepackage[font=small,labelfont=bf,hypcap=false]{caption} 
\usepackage{subcaption}
\usepackage{wrapfig}
\usepackage[numbers]{natbib}

\usepackage{hyperref}

\usepackage{enumitem}
\setlist{nolistsep}

\usepackage[final]{neurips_2020}

\usepackage{array}
\newcolumntype{L}[1]{>{\raggedright\let\newline\\\arraybackslash\hspace{0pt}}m{#1}}
\newcolumntype{C}[1]{>{\centering\let\newline\\\arraybackslash\hspace{0pt}}m{#1}}
\newcolumntype{R}[1]{>{\raggedleft\let\newline\\\arraybackslash\hspace{0pt}}m{#1}}

\title{Value Function Based Performance Optimization\\of Deep Learning Workloads}

\author{
  Benoit Steiner  \\
  Facebook AI Research \\
  \texttt{benoitsteiner@fb.com} \\
  \AND
  Chris Cummins \\
  Facebook AI Research \\
  \texttt{cummins@fb.com} \\
  \And
  Horace He \\
  Facebook \\
  \texttt{chilli@fb.com} \\
  \And
  Hugh Leather \\
  Facebook AI Research \\
  \texttt{hleather@fb.com} \\
}

\begin{document}

\maketitle

\vskip 0.3in

\begin{abstract}

As machine learning techniques become ubiquitous, the efficiency of neural network implementations is becoming correspondingly paramount. Frameworks, such as Halide and TVM, separate out the algorithmic representation of the network from the schedule that determines its implementation. Finding good schedules, however, remains extremely challenging. We model this scheduling problem as a sequence of optimization choices, and present a new technique to accurately predict the expected performance of a partial schedule. By leveraging these predictions we can make these optimization decisions greedily and rapidly identify an efficient schedule. This enables us to find schedules that improve the throughput of deep neural networks by $2.6\times$ over Halide and $1.5\times$ over TVM. Moreover, our technique is two to three orders of magnitude faster than that of these tools, and completes in seconds instead of hours.

\end{abstract}

\section{Introduction}
\label{introduction}

   The rise of deep learning has been accompanied by the development of frameworks such as PyTorch~\citep{pytorch} and TensorFlow~\citep{TF}. The majority of these tools provide a set of primitive tensor operators to perform tasks such as matrix multiplication, convolution and pooling, which are applied in sequence by a runtime interpreter to derive the outputs of the neural network from its input tensors.

  Though pervasive, this approach has two main downsides. First, it requires the development, optimization, and maintenance of a large set of operators, which necessitates scarce human expertise.  As a result, frameworks tend to focus on the most common operators which are only optimized for a limited set of use cases, leaving a lot of performance on the table. Second, these operators can only exchange data through global memory. This is a significant bottleneck, especially in the case of operators of low arithmetic intensity such as activation functions.
  
  To avoid these problems, projects such as Halide~\citep{halide} and TVM~\citep{tvm_compiler} proposed to represent tensor computations using a declarative domain specific language based on Einstein's notation. This high-level representation is then compiled into assembly code that can be executed directly on hardware. This approach abstracts away key implementation choices such as loop ordering, blocking, vectorization, unrolling, inlining, or parallelization, and leaves it up to the compiler to figure out which solution, a.k.a. schedule, most efficiently leverages the available hardware resources. 
  
  Previous work \citep{halide_autoschedule, tvm, flextensor, Ansor20, Sioutas18} has attempted to tackle scheduling by framing the problem as a search in the space of valid implementations. However, these approaches rely on an extensive exploration of the optimization landscape, coupled with aggressive pruning of the solution space. Nevertheless, given the combinatorial nature of the problem, they take several hours to identify a solution and often end up generating suboptimal code.

  In this work, we propose a method to overcome these  limitations. We iteratively improve a value function capable of predicting, given a partial set of scheduling decisions, the best achievable performance over all remaining decisions. This ``look-ahead'' allows us to incrementally schedule an entire neural network by making a set of local decisions that are globally optimal. We show that our technique is able to identify schedules which are on average $1.5\times$ and $2.6\times$ faster than TVM and Halide respectively. We also demonstrate that our approach identifies solutions in two or more orders of magnitude less time.

\section{Automated Scheduling}
\label{scheduling}

\begin{table*}
\vspace{-7mm}
 \centering
 \footnotesize
 \begin{tabular}{c L{8cm}} 
 \toprule
 \textbf{Optimization}  & \textbf{Description} \\
 \midrule
 \texttt{split} & Transform a loop into two nested loops. \\ 
 \texttt{reorder} &  Exchange two loops.  \\
 \texttt{vectorize} & Use SIMD instructions to encode the loop.  \\
 \texttt{parallel} & Parallelize the computation over multiple CPU cores.\\
 \texttt{compute\_at} & Inline the evaluation of a loop into another one. \\
 \texttt{store\_at} & Store the values generated by a layer into a temporary buffer.  \\
 \bottomrule
\end{tabular}
\caption{Primitive scheduling actions used to optimize each layer of a neural network.}
 \label{table:ways-to-optimize}
\end{table*}

\begin{wrapfigure}{r}{0.38\textwidth}
\vspace{-8mm}
\begin{minipage}{0.38\textwidth}

\begin{algorithm}[H]
\begin{algorithmic}
   \STATE {\bfseries Input1:} NN with n layers $L_1$ ... $L_n$
   \STATE {\bfseries Input2:} Value function $V(s)$

   \STATE $s_0$ = InitialState
   \FOR{$i=1$ {\bfseries to} $n$}
      \STATE $C_i$ = CandidateActions($L_i$, $s_{i-1}$)
      \STATE $v_i$ = $\infty$
      \FOR{$s$ {\bfseries in} $C_i$}
        \IF{$V(s) < v_i$}
         \STATE $v_i$ = V($s$)
         \STATE $s_i$ = $s$
        \ENDIF
      \ENDFOR
   \ENDFOR
   
   \STATE {\bfseries Return:} $s_1$ ... $s_n$
\end{algorithmic}
\caption{Scheduling}
 \label{alg:mdp}
\end{algorithm}
\end{minipage}
\vspace{-7mm}
\end{wrapfigure}

We model the problem of choosing the best schedule amongst all the possible options as a deterministic Markov Decision Process (MDP) over a finite horizon with a dynamic action space. In our formulation, a state $s_i$ is a partial schedule -- a set of scheduling decisions applied to the first $i$ layers $L_i$ of a neural network. The set of actions $a_i$ available in state $s_i$ is the set of valid scheduling options available for layer $L_{i+1}$ given the loop structure for the previous layers imposed by the schedule $s_i$. The set of candidate actions we consider is listed in Table~\ref{table:ways-to-optimize}.

We solve the MDP by learning an approximation $V(s)$ of the optimal value function $V^{*}(s)$. In layman's terms, $V^{*}(s)$ is a function capable of predicting the lowest runtime achievable from a state $s$ assuming that we make optimal scheduling decisions for all the subsequent layers of the neural network. Once we have our value function approximation $V(s)$, we greedily schedule the neural network layer by layer following the steps outlined in Algorithm~\ref{alg:mdp}.

For a $N$-layer neural network, with an average of $M$ choices available per layer, we only need to consider $N$*$M$ candidates out of the $M^N$ available complete schedules. This enables us to schedule deep learning workloads extremely quickly as we will see in Section~\ref{benchmarks}.

Note that if we could learn the true value function $V^{*}(s)$ instead of an approximation $V(s)$ our approach would ensure that we find the optimum solution. We'll see in section ~\ref{value_function_evaluation} how each iteration improve the performance of the schedules identified by our approach.
\section{Value Function Estimation}
\label{value_function}

\subsection{Iterative Approximation}

Our technique is inspired from the well known value iteration approach summarized in~\cite{rl_intro}. However, we cannot use this approach directly. Indeed, we face two main hurdles. First, obtaining the reward is very expensive -- it requires compilation of our schedule as well as benchmarking the schedule on actual hardware. Second, it would be impractical to exhaustively visit all the states associated with the scheduling of a single neural network for all but the smallest networks.

\begin{wrapfigure}{r}{0.5\textwidth}
\vspace{-7mm}
\begin{minipage}{0.5\textwidth}
\begin{algorithm}[H]
\begin{algorithmic}
  \STATE {\bfseries Input:} set of neural networks N
  \STATE {\bfseries Input:} value function $V_{i-1}(s)$

  \STATE Initialize $V_i(s)$ to $V_{i-1}(s)$
  \FOR{$n$ {\bfseries in} $N$}
      \FOR{$k$ {\bfseries in} $[0, 100]$}
      \STATE $s_0, ... s_n = \text{BestSchedule}(n, V_{i, \epsilon_k})$
      \FOR{$s_j$ {\bfseries in} $s_0, ... s_n$}
        \STATE $t_{j+1}, ... t_{n} = \text{BeamSearch}(s_j, V_{i-1}))$
        \STATE $r = \text{Benchmark}(s_0, ... s_j, t_{j+1}, ... t_n)$
        \STATE $V_i(s_j) = \min(r, V_i(s_j))$
      \ENDFOR
      \ENDFOR
  \ENDFOR

  \STATE {\bfseries Return:} $V_i(s)$

 \end{algorithmic}
\caption{Single iteration of our value function approximation}
 \label{alg:value_approximation}
\end{algorithm}
\end{minipage}
\vspace{-7mm}
\end{wrapfigure}

On the other hand, we can make a few simplifications. First, we do not need to uniformly sample all the actions available from a state $s$. Although we need a precise estimate of the value function associated with the ``best`` states, we only need a rough estimate for the less interesting states. Consequently, we can sample less often the actions that lead to the less interesting states. Moreover, we can derive an upper bound for the value function for a state $s$ by searching for the best schedule starting from $s$.

Based on these observations, we designed an iterative algorithm, that, starting from an initial approximation of the value function $V_0$, builds progressively more accurate estimates $V_i(s)$.

Algorithm~\ref{alg:value_approximation} details how we perform each iteration: we extract 100 schedules for each neural network in our training set. We inject a small amount of random noise $\epsilon$ to the predictions made by the value function to ensure that we cover a significant portion of the interesting states of each pipeline. Starting from each state $s_i$ in the previously identified schedules, we run a beam search guided by our previous estimate of the value function, and benchmark the resulting schedule. We use the measured runtime to refine the estimate of the value function V for the state $s_i$. To bootstrap the process, we modify algorithm~\ref{alg:value_approximation} to generate and benchmark end to end schedules purely randomly. 

\subsection{Implementation}

The throughput of a neural network depends on two main factors: the amount of computation and data access to be performed, and the overall organization of the computation. Consequently, we devised two groups of input features to capture this information: a set of intrinsic features that are invariant to the schedule, and a collection of schedule dependant features that are acquired as the process of scheduling a pipeline progresses. In this setting, the intrinsic features enable our model to predict how fast each stage could be executed if scheduled optimally, while the acquired features enable us to capture how well the scheduled stages are expected to perform. 

Since the set of scheduling decisions made for the initial layers of a neural network impact the performance of the yet to be scheduled layers, we architected our value function around an LSTM, that we feed with the normalized values of our intrinsic and acquired features. To predict the expected performance of the whole neural network, we sum the predictions of each of the timesteps of the LSTM.

\section{Evaluation}

\begin{wrapfigure}{r}{0.5\textwidth}
\vspace{-13mm}
\includegraphics[width=0.5\textwidth]{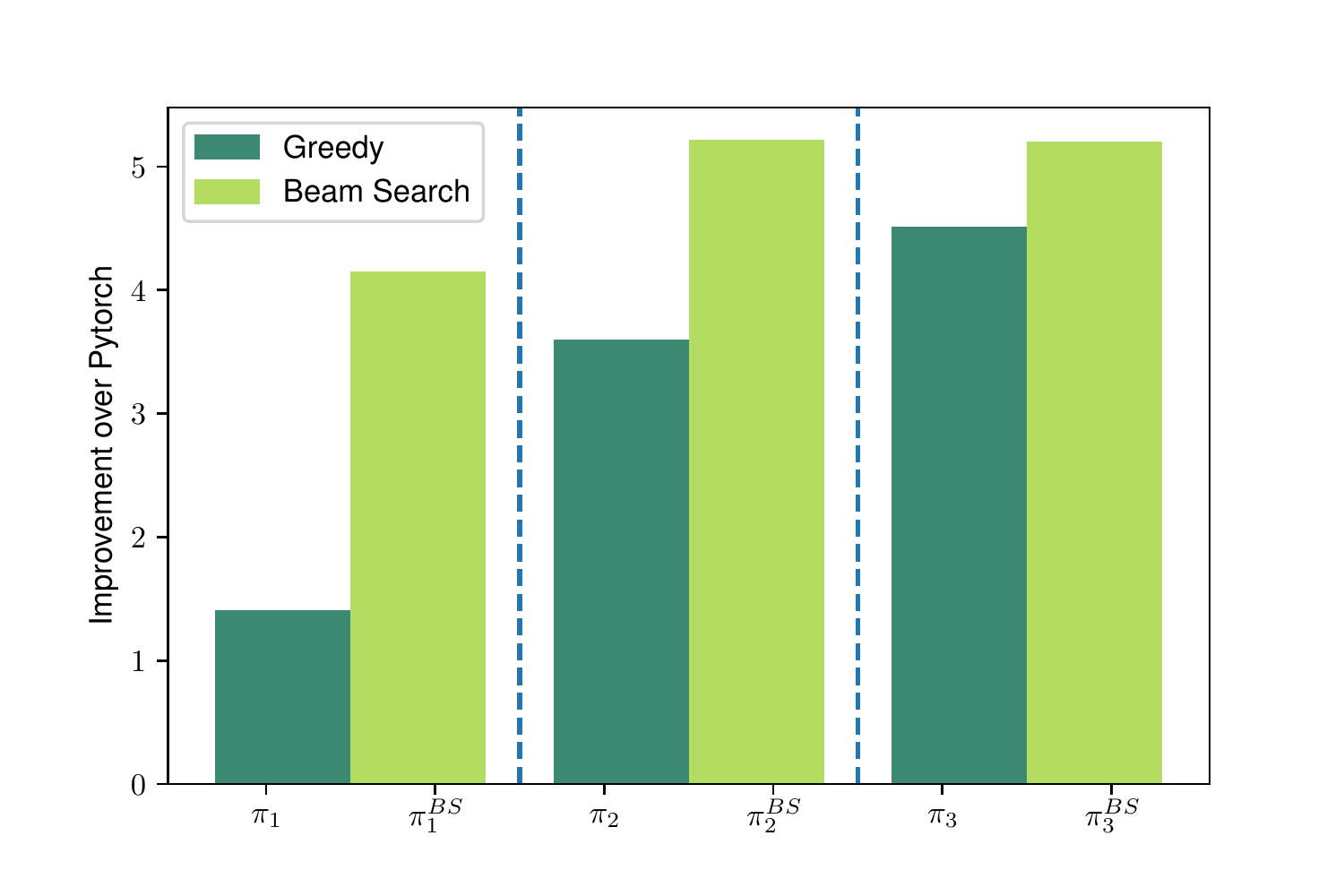}
\caption{Improvement in our value function through 2 rounds of value improvement. Each round significantly improves the quality of our generated schedules.}
\label{fig:value_function_improvements}

\vspace{-20mm}
\end{wrapfigure}

We ran all our experiments on an Intel Xeon D-2191A CPU running at 1.60GHz with 48GB of RAM and a NVidia Tesla M40 GPU. We used Halide~\cite{halide} to compile our schedules into assembly code.

\subsection{Value Function}
\label{value_function_evaluation}

First, we measured the impact of the inductive bias inherent in our model architecture on its ability to encode the value function. Across our rounds of value improvement $V_i$, our neural network was able to predict the expected values with an average error inferior to 5\% and an $R^2$ greater than 0.955. 

Next, we would like to demonstrate that our iterative value learning process is effective. However, we cannot show that our successive value function approximations $V_i(s)$ converge towards the optimal value function $V^{*}(s)$ since determining the exact value of $V^{*}(s)$ would require an exhaustive search in a combinatorially large solution space. Instead we show that with each iteration our value function estimates are better able to guide a search. In Figure \ref{fig:value_function_improvements}, we plot the relative performance of the schedules selected by our greedy search as well as a standard beam search under the guidance of three successive estimates V0, V1, and V2 on the 12 models used in Figure~\ref{figure:results}. As V0, V1, and V2 refine the estimates of our value function the gap between the quality of the schedules identified by a greedy and a beam search decreases, while the overall performance of the schedules increases

\subsection{Benchmarks}
\label{benchmarks}

\begin{figure*}[b]
  \begin{subfigure}[t]{\textwidth}
    \includegraphics[width=\linewidth]{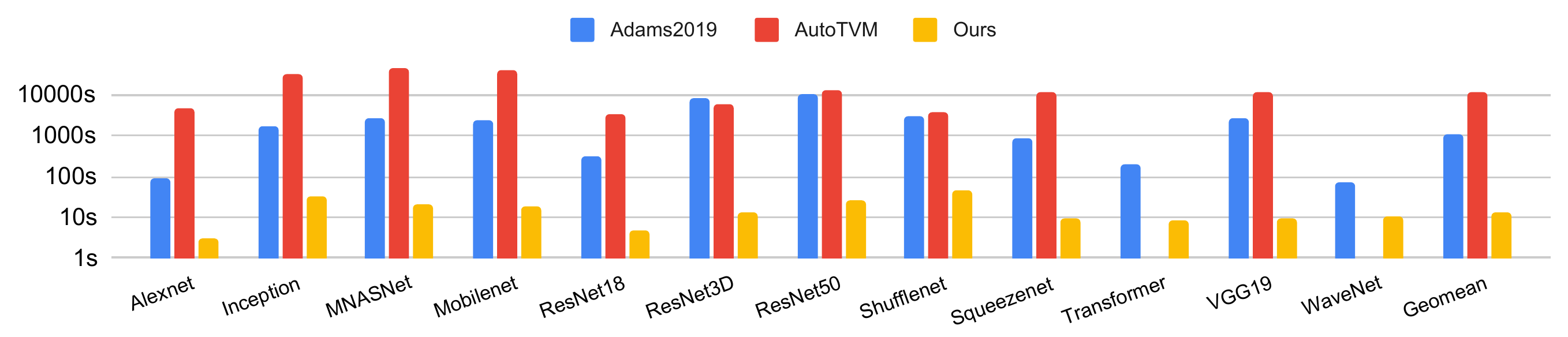}
    \vspace{-.3cm}
    \caption{Schedule search time. Lower is better.}
    \label{fig:search_time}
  \end{subfigure}
    \\*
  \begin{subfigure}[t]{\textwidth}
    \centering%
    \includegraphics[width=\linewidth]{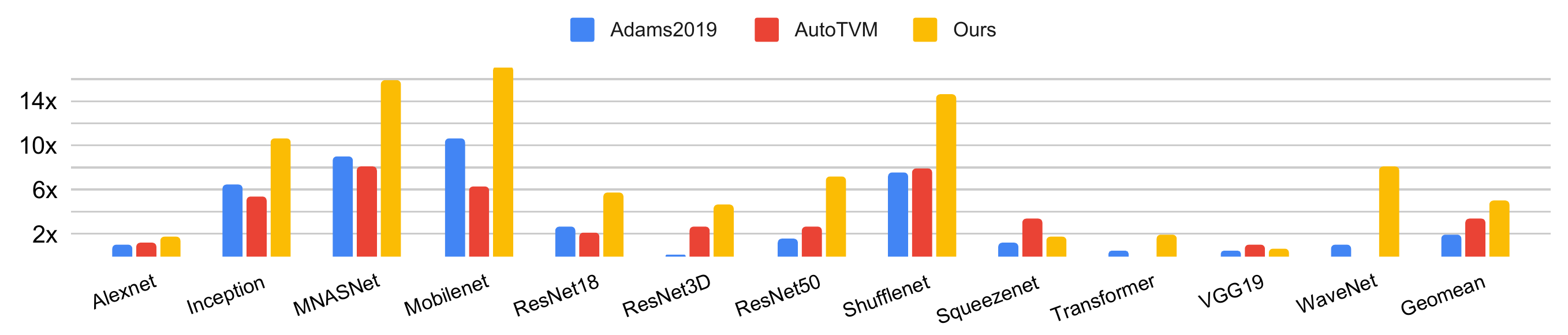}
    \vspace{-.3cm}
    \caption{Speedup of schedules over PyTorch. Higher is better.}
    \label{figure:schedule-speedup}
  \end{subfigure}

  \caption{
     (\subref{fig:search_time}) Time taken to schedule deep learning workloads, and (\subref{figure:schedule-speedup}) performance of final schedules relative to PyTorch. We plot the search times on a log scale. We compare our results against the Halide autoscheduler (Adams2019) and AutoTVM using the published configuration. AutoTVM failed to load the Transformer and Wavenet models.
  }
\label{figure:results}
\end{figure*}

We evaluate our search strategy on a diverse set of deep learning workloads encompassing text, speech, and image processing tasks.

In Figure~\ref{figure:results} we compare our implementation against the following systems: PyTorch 1.6, AutoTVM version 0.6, and the Halide auto-scheduler version 8.0.0. To run AutoTVM, we followed the steps outlined in its documentation~\citep{tvm-autotune}. We ran the Halide auto-scheduler with its default settings of 5 search passes with each pass identifying 32 candidate schedule, and benchmarking to do the final ranking of the 160 candidates.

First, we show the search time of the systems in Figure~\ref{fig:search_time}. PyTorch is extremely fast, since it does not search for good solutions. AutoTVM takes considerable time, requiring 3 hours on average and up to 12 hours to complete the search. Halide takes an average of 20 minutes and up to 2.5 hours. Our approach schedules the neural networks in 13 seconds on average, and 47 seconds in the worst case.

However, this significantly faster search time does not come at the cost of lower quality schedules, as can be seen in Figure~\ref{figure:schedule-speedup}. Our search finds schedules that outperform PyTorch by a factor of $5.1\times$. We also improve on AutoTVM and Halide with a speedup of $1.5\times$ and $2.6\times$ respectively.

\section{Conclusion}
Our results demonstrate that automatically learning complex neural network scheduling policies using reinforcement learning is feasible and lead to better results in a fraction of the search time. We hope that this will allow these techniques to be used in a broader range of applications.

\bibliographystyle{unsrt}
\bibliography{12-references.tex}

\begin{thebibliography}{10}

\bibitem{pytorch}
Adam Paszke, Sam Gross, Francisco Massa, Adam Lerer, James Bradbury, Gregory
  Chanan, Trevor Killeen, Zeming Lin, Natalia Gimelshein, Luca Antiga, Alban
  Desmaison, Andreas Kopf, Edward Yang, Zachary DeVito, Martin Raison, Alykhan
  Tejani, Sasank Chilamkurthy, Benoit Steiner, Lu~Fang, Junjie Bai, and Soumith
  Chintala.
\newblock Pytorch: An imperative style, high-performance deep learning library.
\newblock In {\em Advances in Neural Information Processing Systems 32}, pages
  8026--8037. Curran Associates, Inc., 2019.

\bibitem{TF}
Mart\'{\i}n Abadi, Ashish Agarwal, Paul Barham, Eugene Brevdo, Zhifeng Chen,
  Craig Citro, Greg~S. Corrado, Andy Davis, Jeffrey Dean, Matthieu Devin,
  Sanjay Ghemawat, Ian Goodfellow, Andrew Harp, Geoffrey Irving, Michael Isard,
  Yangqing Jia, Rafal Jozefowicz, Lukasz Kaiser, Manjunath Kudlur, Josh
  Levenberg, Dandelion Man\'{e}, Rajat Monga, Sherry Moore, Derek Murray, Chris
  Olah, Mike Schuster, Jonathon Shlens, Benoit Steiner, Ilya Sutskever, Kunal
  Talwar, Paul Tucker, Vincent Vanhoucke, Vijay Vasudevan, Fernanda Vi\'{e}gas,
  Oriol Vinyals, Pete Warden, Martin Wattenberg, Martin Wicke, Yuan Yu, and
  Xiaoqiang Zheng.
\newblock {TensorFlow}: Large-scale machine learning on heterogeneous systems,
  2015.
\newblock Software available from tensorflow.org.

\bibitem{halide}
Jonathan Ragan-Kelley, Andrew Adams, Sylvain Paris, Marc Levoy, Saman
  Amarasinghe, and Frédo Durand.
\newblock Decoupling algorithms from schedules for easy optimization of image
  processing pipelines.
\newblock {\em ACM Transactions on Graphics - TOG}, 31, 07 2012.

\bibitem{tvm_compiler}
Tianqi Chen, Thierry Moreau, Ziheng Jiang, Lianmin Zheng, Eddie Yan, Haichen
  Shen, Meghan Cowan, Leyuan Wang, Yuwei Hu, Luis Ceze, Carlos Guestrin, and
  Arvind Krishnamurthy.
\newblock {TVM}: An automated end-to-end optimizing compiler for deep learning.
\newblock In {\em 13th {USENIX} Symposium on Operating Systems Design and
  Implementation ({OSDI} 18)}, pages 578--594, Carlsbad, CA, October 2018.
  {USENIX} Association.

\bibitem{halide_autoschedule}
Andrew Adams, Karima Ma, Luke Anderson, Riyadh Baghdadi, Tzu-Mao Li,
  Micha\"{e}l Gharbi, Benoit Steiner, Steven Johnson, Kayvon Fatahalian,
  Fr{\'e}do Durand, and Jonathan Ragan-Kelley.
\newblock Learning to optimize halide with tree search and random programs.
\newblock {\em ACM Trans. Graph.}, 38(4):121:1--121:12, July 2019.

\bibitem{tvm}
Tianqi Chen, Lianmin Zheng, Eddie Yan, Ziheng Jiang, Thierry Moreau, Luis Ceze,
  Carlos Guestrin, and Arvind Krishnamurthy.
\newblock Learning to optimize tensor programs.
\newblock In {\em Proceedings of the 32Nd International Conference on Neural
  Information Processing Systems}, NIPS'18, pages 3393--3404, USA, 2018. Curran
  Associates Inc.

\bibitem{flextensor}
Size Zheng, Yun Liang, Shuo Wang, Renze Chen, and Kaiwen Sheng.
\newblock Flextensor: An automatic schedule exploration and optimization
  framework for tensor computation on heterogeneous system.
\newblock ASPLOS '20, page 859–873, New York, NY, USA, 2020. Association for
  Computing Machinery.

\bibitem{Ansor20}
Lianmin Zheng, Chengfan Jia, Minmin Sun, Zhao Wu, Cody~Hao Yu, Ameer Haj-Ali,
  Yida Wang, Jun Yang, Danyang Zhuo, Koushik Sen, Joseph~E. Gonzalez, and Ion
  Stoica.
\newblock Ansor: Generating high-performance tensor programs for deep learning.
\newblock In {\em 14th {USENIX} Symposium on Operating Systems Design and
  Implementation ({OSDI} 20)}, Banff, Alberta, November 2020. {USENIX}
  Association.

\bibitem{Sioutas18}
Savvas Sioutas, Sander Stuijk, Henk Corporaal, Twan Basten, and Lou Somers.
\newblock Loop transformations leveraging hardware prefetching.
\newblock In {\em Proceedings of the 2018 International Symposium on Code
  Generation and Optimization}, CGO 2018, page 254–264, New York, NY, USA,
  2018. Association for Computing Machinery.

\bibitem{rl_intro}
Richard~S. Sutton and Andrew~G. Barto.
\newblock {\em Reinforcement Learning: An Introduction}.
\newblock The MIT Press, second edition, 2018.

\bibitem{tvm-autotune}
Yao Wang and Eddie Yan.
\newblock Auto-tuning a convolutional network for x86 cpu.
\newblock
  \url{https://tvm.apache.org/docs/tutorials/autotvm/tune_relay_x86.html}.
\newblock Accessed: 2020-09-30.

\end{thebibliography}

\end{document}